\documentclass{article}

\usepackage{arxiv}

\usepackage[utf8]{inputenc} 
\usepackage[T1]{fontenc}    
\usepackage{hyperref}       
\usepackage{url}            
\usepackage{booktabs}       
\usepackage{amsfonts}       
\usepackage{nicefrac}       
\usepackage{microtype}      
\usepackage{lipsum}

\usepackage{bm}
\usepackage{amsmath}
\usepackage{amssymb}
\usepackage{amsfonts}
\usepackage{graphicx}
\def\vec#1{\mathbf{#1}}
\usepackage{algorithm}
\usepackage{algorithmic}
\usepackage{comment}
\newcommand{\argmax}{\mathop{\rm arg~max}\limits}

\title{Meta-learning for Out-of-Distribution Detection via Density Estimation in Latent Space}

\author{
  Tomoharu Iwata\\
  NTT Communication Science Laboratories\\
  \AND
  Atsutoshi Kumagai\\
  NTT Computer and Data Science Laboratories
}
\date{}

\begin{document}
\maketitle
\begin{abstract}
  Many neural network-based out-of-distribution (OoD) detection methods have been proposed. However, they require many training data for each target task. We propose a simple yet effective meta-learning method to detect OoD with small in-distribution data in a target task. With the proposed method, the OoD detection is performed by density estimation in a latent space. A neural network shared among all tasks is used to flexibly map instances in the original space to the latent space. The neural network is meta-learned such that the expected OoD detection performance is improved by using various tasks that are different from the target tasks. This meta-learning procedure enables us to obtain appropriate representations in the latent space for OoD detection. For density estimation, we use a Gaussian mixture model (GMM) with full covariance for each class. We can adapt the GMM parameters to in-distribution data in each task in a closed form by maximizing the likelihood. Since the closed form solution is differentiable, we can meta-learn the neural network efficiently with a stochastic gradient descent method by incorporating the solution into the meta-learning objective function. In experiments using six datasets, we demonstrate that the proposed method achieves better performance than existing meta-learning and OoD detection methods.
  \end{abstract}

\section{Introduction}

Out-of-Distribution (OoD) detection is an important machine learning
problem that finds instances that do not belong to training classes~\cite{liang2018enhancing}. 
Deep learning models tend to make incorrect predictions
with high confidence for instances from unseen
classes~\cite{szegedy2014intriguing,moosavi2017universal}.
By OoD detection, we can safely deploy machine learning models
in the open world, where new classes can emerge.
OoD detection can also be beneficial for 
ensuring the quality of the collected data,
and finding instances with unusual behavior for data analyists~\cite{yu2019unsupervised}.
Many neural network-based OoD detection methods have been proposed~\cite{morningstar2021density,ren2019likelihood,devries2018learning,winkens2020contrastive,liu2020energy,chen2020robust,xiao2020likelihood,feng2021improving}.
However, these methods require a large number of data for training.
In some real-world applications, enough data might be unavailable
in target tasks.
For example, it is important to detect manufacturing failures
that are not categorized in existing failure classes in each factory
for improving productivity,
but many data are not accumulated in newly operated factories.

We propose a simple yet effective method
for improving the OoD detection performance on unseen target tasks
by meta-learning from data in tasks different from the target tasks.
Even when enough data are unavailable for a newly operated factory,
many data for different factories would be available.
For the meta-training data,
instances with class labels from various tasks are assumed to be given,
where classes are different across tasks.
For a target task,
a small number of instances with class labels
are given as in-distribution (ID) data.
We want to identify whether test instances
are ID or OoD for the target task.

With the proposed method,
OoD scores are calculated by density estimation in a latent space.
Density estimation has been used for OoD detection~\cite{bishop1994novelty}.
However, ID data can have low likelihoods 
in extremely high dimensions, and density estimation
in the original high dimensional space sometimes fails
to detect OoD~\cite{choi2018waic,nalisnick2018deep,hendrycks2018deep}.
To avoid such problems,
the proposed method embeds instances
into a latent space by a neural network that is shared among all tasks.

Meta-learning is formulated as a bilevel optimization problem.
In the inner optimization problem,
task-specific parameters are adapted to the given task-specific data.
In the outer optimization problem,
common parameters shared across all tasks
are meta-learned to improve the expected test performance
when the task-specific parameters
adapted in the inner optimization are used.
With the proposed method,
the inner optimization problem
corresponds to fitting a density model
to the task-specific data in a latent space.
We use a Gaussian mixture model (GMM) for the density model.
The GMM parameters,
which are the class probability, mean, and covariance for each class,
are adapted in a closed form by maximizing the likelihood.
In the outer optimization problem,
we meta-learns the neural network
such that the OoD detection performance improves
when the density is estimated with the adapted GMM parameters.
By incorporating the closed form solution of the inner optimization problem
into the outer optimization problem,
we can solve the bilevel optimization problem efficiently
by a stochastic gradient descent method.

\begin{figure}[t!]
  \centering
  \includegraphics[width=21em]{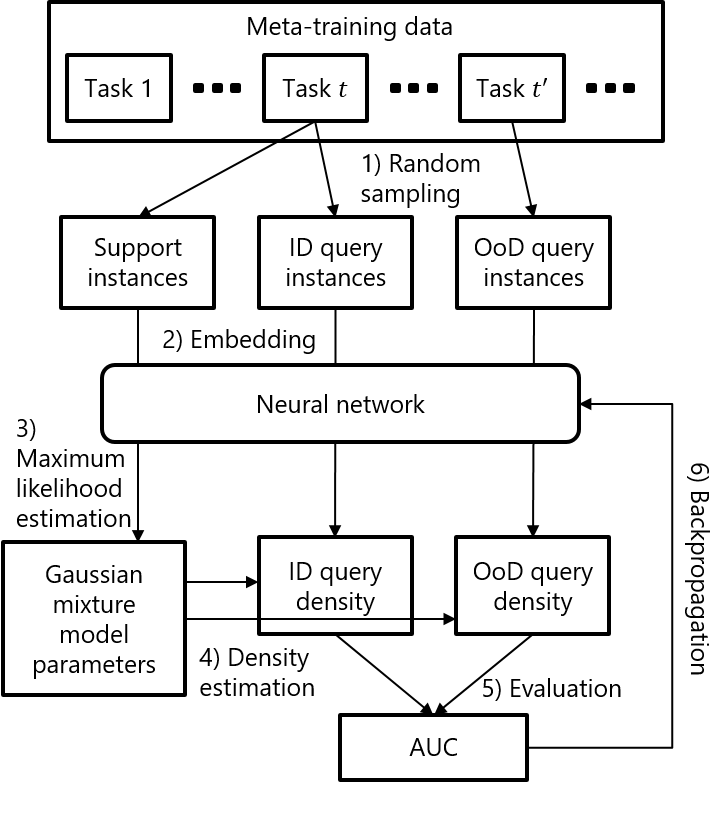}
  \caption{Meta-learning step for the proposed method. For the meta-training data, we are given multiple tasks with different classes. 1) Randomly sample support instances and ID query instances from a task,
    and sample OoD query instances from another task.
    2) Embed sampled instances by a neural network in a latent space.
    3) Estimate task-specific Gaussian mixture model parameters based on the maximum likelihood.
    4) Calculate the density of the query instances using the estimated GMM parameters.
    5) Evaluate the OoD score based on AUC.
    6) Backpropagate the AUC to update the neural network parameters.}
  \label{fig:model}
\end{figure}

For each meta-learning step,
instances in the meta-training data
are randomly selected for ID,
and instances in classes different from the ID instances
are randomly selected for OoD.
By generating ID and OoD data in this way,
we can evaluate the OoD detection performance using the meta-training data,
which is required for the objective function
in the outer optimization problem.
The performance is evaluated by the area under the ROC curve (AUC),
which is often used in
the existing literature~\cite{liang2018enhancing,wang2019out,yu2019unsupervised}.
Although data in target tasks are labeled with their classes,
only ID instances are given as the task-specific data.
Therefore, the inner optimization needs to be 
performed without OoD instances.
The proposed method enables this by
defining the inner optimization as density estimation using ID instances. 
Figure~\ref{fig:model} illustrates
a meta-learning step of the proposed method.

The main contributions of this paper are as follows:
\begin{enumerate}
\item We propose a meta-learning method for OoD detection that uses labeled data in various tasks to improve the performance on unseen target tasks that consist of unlabeled data.
\item The proposed method enables an efficient bilevel optimization by modeling the inner optimization with the maximum likelihood of a Gaussian mixture model in a latent space, which gives a closed-form solution for the adaptation to each task.
\item We confirm that the proposed method achieves better performance than existing methods.
\end{enumerate}
The remainder of this paper is organized as follows.
In Section~\ref{sec:related},
we briefly review related work.
In Section~\ref{sec:proposed},
we formulate our problem, 
and present our meta-learning method for OoD detection.
In Section~\ref{sec:experiments},
we demonstrate that the proposed method achieves better performance than existing meta-learning and OoD detection methods.
Finally, we present concluding remarks and a discussion of future work
in Section~\ref{sec:conclusion}.

\section{Related work}
\label{sec:related}

Many meta-learning methods have been proposed~\cite{schmidhuber:1987:srl,bengio1991learning,finn2017model,vinyals2016matching,snell2017prototypical,lee2019meta}.
Meta-learning methods learn how to learn from a small amount of
labeled data in various tasks,
and use the learned knowledge in unseen tasks.
However, most of these methods are not designed for OoD detection,
and are inapplicable to our problem.
A meta-learning method for OoD detection
based on model-agnostic meta-learning (MAML)~\cite{finn2017model}
has been proposed~\cite{jeong2020ood}.
For the inner optimization,
MAML requires costly back-propagation through iterative gradient descent steps.
On the other hand, the proposed method achieves an efficient inner optimization
in a closed form based on density estimation with GMMs.
The closed form inner optimization for meta-learning has been successfully used
in various methods~\cite{bertinetto2018meta,iwata2020few,snell2017prototypical,fortuin2019deep,harrison2018meta}.
However, they define the inner optimization problem as
regression or classification, but not as density estimation.
An OoD detection method based on Mahalanobis distance 
assumes a GMM in a latent space~\cite{lee2018simple}.
However, it uses a covariance shared among all classes,
which has lower expressive power than class-specific covariances.
In addition,
it requires many training data since it is not a meta-learning method.
Anomaly detection is related to OoD detection.
Anomaly detection methods usually do not use
class labels~\cite{chalapathy2018anomaly,kumagai2019transfer,akcay2018ganomaly,zenati2018efficient,schlegl2017unsupervised},
where class labels indicate categories that ID instances belong to
and they are different from ID/OoD labels.
On the other hand, the proposed method uses class labels
for effectively estimating the density.
The better performance of the proposed method
compared with these existing methods are demonstrated in our experiments
in Section~\ref{sec:experiments}.

\section{Proposed method}
\label{sec:proposed}

\subsection{Problem formulation}

In a meta-training phase,
meta-training data
$\mathcal{D}=\{\{(\vec{x}_{tn},y_{tn})\}_{n=1}^{N_{t}}\}_{t=1}^{T}$ are given,
where $T$ is the number of tasks,
$N_{t}$ is the number of labeled instances in the $t$th task,
$\vec{x}_{tn}\in\mathcal{X}$ is the feature of the $n$th instance,
$y_{tn}\in\{c_{t1},\dots,c_{tK_{t}}\}$ is its class label,
$c_{tk}$ is the $k$th class,
and $K_{t}$ is the number of classes.
In a meta-test phase,
labeled support set $\mathcal{S}=\{(\vec{x}_{n},y_{n})\}_{n=1}^{N_{\mathrm{S}}}$
is given in a target task that are different from the training tasks,
where $y_{n}\in\{c_{1},\cdots,c_{K}\}$,
$K$ is the number of classes in the support set,
and the classes in the target task do not overlap with
those in the training tasks.
Our aim is to identify
whether unlabeled instances $\{\vec{x}\}$ 
belong to the classes
in the support set $\{c_{1},\cdots,c_{K}\}$ (in-distribution)
or not (out-of-distribution).
We assume that feature space $\mathcal{X}$ is the same across all tasks.

\subsection{Model}

In this subsection,
we present our model to output
task-specific OoD score $u(\vec{x}\mid\mathcal{S};\bm{\Phi})$
of unlabeled instance $\vec{x}$
given supposet set
$\mathcal{S}=\{(\vec{x}_{n},y_{n})\}_{n=1}^{N_{\mathrm{S}}}$,
where $\bm{\Phi}$ is the set of the common model parameters shared across tasks.

The proposed model
uses the following negative log density in a latent space
for the OoD score,
\begin{equation}
  u(\vec{x}\mid\mathcal{S};\bm{\Phi}) = -\log p\left(f(\vec{x};\bm{\phi})\mid\hat{\bm{\Theta}}(\mathcal{S};\bm{\Phi})\right),
  \label{eq:u}
\end{equation}
where $p$ is the probabilisty density function in the latent space,
$f:\mathcal{X}\rightarrow\mathbb{R}^{D}$
is a neural network that maps an instance from
the original space to the $D$-dimensional latent space,
$\bm{\phi}\in\bm{\Phi}$ is the neural network parameters
that are shared across tasks,
and
$\hat{\bm{\Theta}}(\mathcal{S};\bm{\Phi})$ is the set of
the task-specific parameters
of the probability density function adapted to support set $\mathcal{S}$.
The negative log likelihood is an natural OoD score
since instances with the low probability density can be seen OoD,
and it has been used for OoD detection~\cite{bishop1994novelty}.
With neural network $f$,
we can find a latent space that is appropriate for detecting OoD
even when the density estimation in the original space
does not perform well on OoD detection.

The probability density function parameters $\bm{\Theta}$
is adapted to support set $\mathcal{S}$ by maximizing the log likelihood,
\begin{align}
  \hat{\bm{\Theta}}(\mathcal{S};\bm{\Phi})
  &=\argmax_{\bm{\Theta}}\log p(\mathcal{S}\mid\bm{\Theta};\bm{\Phi})
  \nonumber\\
  &=\argmax_{\bm{\Theta}}\sum_{(\vec{x},y)\in\mathcal{S}}\log p(\vec{x}\mid y,\bm{\Theta};\bm{\Phi}).
  \label{eq:ml}
\end{align}
The proposed model assumes a Gaussian mixture model (GMM)
in the latent space,
\begin{align}
  \lefteqn{p\left(f(\vec{x};\bm{\phi})\mid\bm{\Theta}\right)
    = \sum_{k=1}^{K}\gamma_{k}\mathcal{N}\left(f(\vec{x};\bm{\phi})\mid
    \bm{\mu}_{k},\bm{\Sigma}_{k}\right)}
    \nonumber\\
    &=\sum_{k=1}^{K}\gamma_{k}(2\pi)^{-\frac{D}{2}}\vert\bm{\Sigma}_{k}\vert^{-\frac{1}{2}}
    \exp\left(-\frac{1}{2}(f(\vec{x};\bm{\phi})-\bm{\mu}_{k})^{\top}\bm{\Sigma}_{k}^{-1}(f(\vec{x};\bm{\phi})-\bm{\mu}_{k})\right),
    \label{eq:gmm}
\end{align}
where $\gamma_{k}\in [0,1]$ is the class probability of the $k$th class,
$\mathcal{N}(\cdot\mid\bm{\mu},\bm{\Sigma})$ is the Gaussian distribution
with mean $\bm{\mu}\in\mathbb{R}^{D}$ and covariance $\bm{\Sigma}\in\mathbb{R}^{D\times D}$,
and $\bm{\Theta}=\{\gamma_{k},\bm{\mu}_{k},\bm{\Sigma}_{k}\}_{k=1}^{K}$
is the set of the task-specific GMM parameters.
Since the class label is given for each support instance,
we can obtain the adapted parameters in a closed form using the GMM
by solving Eq.~(\ref{eq:ml}) as follows,
\begin{align}
  \hat{\bm{\gamma}}_{k}(\mathcal{S})
  =\frac{1}{\mid\mathcal{S}_{k}\mid}, \quad
  \hat{\bm{\mu}}_{k}(\mathcal{S})=\frac{1}{\vert\mathcal{S}_{k}\vert}\sum_{\vec{x}\in\mathcal{S}_{k}}f(\vec{x};\bm{\phi}),\nonumber
\end{align}
\begin{align}
  \hat{\bm{\Sigma}}_{k}(\mathcal{S})
    =\frac{1}{\vert\mathcal{S}_{k}\vert}
  \left(\sum_{\vec{x}\in\mathcal{S}_{k}}
  (f(\vec{x};\bm{\phi})-\hat{\bm{\mu}}_{k}(\mathcal{S}))
  (f(\vec{x};\bm{\phi})-\hat{\bm{\mu}}_{k}(\mathcal{S}))^{\top}  
  +\beta\vec{I}\right),
  \label{eq:gmm_params}
\end{align}
where $\mathcal{S}_{k}=\{\vec{x}\mid(\vec{x},y)\in\mathcal{S},y=k\}$ is the set of instances with label $k$ in the support set, $\vert\mathcal{S}_{k}\vert$ is its size, $\beta\in\mathbb{R}_{>0}$ is the parameter,
and $\hat{\bm{\Theta}}(\mathcal{S};\bm{\Phi})=\{\hat{\gamma}_{k}(\mathcal{S}),\hat{\bm{\mu}}_{k}(\mathcal{S}),\hat{\bm{\Sigma}}_{k}(\mathcal{S})\}_{k=1}^{K}$.
By parameter $\beta$, the covariance is regularized,
and is guaranteed to be positive definite,
which is necessary to compute the inverse of the covariance stably.
The inverse is required to calculate
the likelihood in Eq.~(\ref{eq:gmm}).

\subsection{Meta-training}

The common parameters to be meta-learned 
are $\bm{\Phi}=\{\bm{\phi},\beta\}$, where $\bm{\phi}$ is
the set of the neural network parameters, and $\beta$ is the
regularization parameter for covariance estimation.
Let $\mathcal{Q}_{\mathrm{I}}=\{\vec{x}^{\mathrm{I}}_{n}\}_{n=1}^{N_{\mathrm{I}}}$ be a query set of instances
that are ID of the support set,
and $\mathcal{Q}_{\mathrm{O}}=\{\vec{x}_{n}^{\mathrm{O}}\}_{n=1}^{N_{\mathrm{O}}}$
be a query set of instances
that are OoD of the support set.
Each meta-training task contains support set $\mathcal{S}$ and
query set $\mathcal{Q}=\{\mathcal{Q}_{\mathrm{I}},\mathcal{Q}_{\mathrm{O}}\}$.
We want to improve the expected AUC over tasks,
\begin{align}
  \hat{\bm{\Phi}}=
  \argmax_{\bm{\Phi}}\mathbb{E}_{t\in\{1,\cdots,T\}}\left[\mathbb{E}_{\mathcal{S},\mathcal{Q}\subset\mathcal{D}_{t}}\left[\mathrm{AUC}\right(\mathcal{Q}\mid\hat{\bm{\Theta}}(\mathcal{S};\bm{\Phi}),\bm{\Phi}\left)\right]\right],
  \label{eq:expected_auc}
\end{align}
where $\mathbb{E}$ represents the expectation,
and
$\mathrm{AUC}(\mathcal{Q}\mid\bm{\Theta},\bm{\Phi})$
is the AUC on query set $\mathcal{Q}$ calculated using
task-specific parameters $\bm{\Theta}$
and common parameters $\bm{\Phi}$,
and $\hat{\bm{\Theta}}(\mathcal{S};\bm{\Phi})$
is a task-specific parameters adapted to support set $\mathcal{S}$
by maximizing the log likelihood in Eq.~(\ref{eq:ml}).
It is a bilevel optimization problem where
Eq.~(\ref{eq:ml}) is the inner problem,
and Eq.~(\ref{eq:expected_auc}) is the outer problem.
Since the inner problem is solved in a closed form
as shown in Eq.~(\ref{eq:gmm_params}),
the bilevel optimization can be efficiently solved
by incorporating the inner solution into the outer problem.
The objective functions of the inner and outer problems
are the likelihood and AUC, respectively, and they are different.
It is because 
while only ID data are given in the inner problem,
which is the same setting with the meta-test phase,
both of the ID and OoD data are given in the outer problem.

The AUC is calculated by the ratio that the OoD score of
the OoD query instances is higher than that of the ID query instances,
\begin{align}
  \mathrm{AUC}(\mathcal{Q}\mid\hat{\bm{\Theta}}(\mathcal{S};\bm{\Phi}),\bm{\Phi})=
    \frac{1}{N_{\mathrm{O}}N_{\mathrm{I}}}
  \sum_{\vec{x}^{\mathrm{O}}\in\mathcal{Q}_{\mathrm{O}}}
  \sum_{\vec{x}^{\mathrm{I}}\in\mathcal{Q}_{\mathrm{I}}}
  I\left(u(\vec{x}^{\mathrm{O}}\mid\mathcal{S};\bm{\Phi})>u(\vec{x}^{\mathrm{I}}\mid\mathcal{S};\bm{\Phi})\right),
\end{align}
where $I$ is the indicator function, i.e., $I(A)=1$ is $A$ is true,
and $I(A)=0$ otherwise.
To make the AUC differentiable,
we use the following smooth approximation of the AUC,
\begin{align}
  \widetilde{\mathrm{AUC}}(\mathcal{Q}\mid\hat{\bm{\Theta}}(\mathcal{S};\bm{\Phi}),\bm{\Phi})=
    \frac{1}{N_{\mathrm{O}}N_{\mathrm{I}}}
  \sum_{\vec{x}^{\mathrm{O}}\in\mathcal{Q}_{\mathrm{O}}}
  \sum_{\vec{x}^{\mathrm{I}}\in\mathcal{Q}_{\mathrm{I}}}
  \sigma\left(u(\vec{x}^{\mathrm{O}}\mid\mathcal{S};\bm{\Phi})-u(\vec{x}^{\mathrm{I}}\mid\mathcal{S};\bm{\Phi})\right),
  \label{eq:smooth_auc}
\end{align}
where $\sigma(x)=\frac{1}{1+\exp(-x)}$ is the sigmoid function
which is used for the smooth approximation of the indicator function~\cite{iwata2019supervised}.

Algorithm~\ref{alg:train} shows the meta-learning procedures of the
proposed model.
The expectation in Eq.~(\ref{eq:expected_auc}) is calculated
by the Monte Carlo method,
where support and query sets are randomly sampled
from the meta-training data in Lines 3--6.

The time complexity to adapt GMM parameters in Eq.~(\ref{eq:gmm_params}) is
$O(KN^{2}D)$, where
$K$ is the number of classes,
$N$ is the number of support instances per class,
and $D$ is the latent space dimensionality.
The complexity to calculate the OoD score is $O(KD^{3})$,
where the complexity of the inverse of the covariance matrix is
the cubic of the matrix size.
When the latent space dimensionality is larger than the number of instances,
$D>N$, we can reduce the complexity to $O(KN^{3})$
using the Woodbury formula~\cite{petersen2008matrix}.
The complexity to calculate the smooth AUC is $O(N_{\mathrm{O}}N_{\mathrm{I}})$
since it is a summation over $N_{\mathrm{O}}$ OoD query instances
and $N_{\mathrm{I}}$ ID query instances.

\begin{algorithm}[t!]
  \caption{Meta-training procedure of our model.}
  \label{alg:train}
  \begin{algorithmic}[1]
    \renewcommand{\algorithmicrequire}{\textbf{Input:}}
    \renewcommand{\algorithmicensure}{\textbf{Output:}}
    \REQUIRE{Meta-training data $\mathcal{D}$, support set size $N_{\mathrm{S}}$, ID query set size $N_{\mathrm{I}}$, OoD query set size $N_{\mathrm{O}}$}
    \ENSURE{Trained common parameters $\bm{\Phi}$}
    \STATE Initialize common parameters $\bm{\Phi}$.
    \WHILE{End condition is satisfied}
    \STATE Randomly select task index $t$ from $\{1,\cdots,T\}$.
    \STATE Generate support set $\mathcal{S}\subset\mathcal{D}_{t}$ by randomly sampling $N_{\mathrm{S}}$ instances from the selected task.
    \STATE Generate ID query set $\mathcal{Q}_{\mathrm{I}}$ by randomly sampling $N_{\mathrm{I}}$ instances from the selected task, $\mathcal{Q}_{\mathrm{I}}\subset\mathcal{D}_{t}\setminus\mathcal{S}$ where the ID query set and support set do not share the instances, $\mathcal{Q}_{\mathrm{I}}\cap\mathcal{S}=\phi$.
    \STATE Generate OoD query set $\mathcal{Q}_{\mathrm{O}}$ by randomly sampling $N_{\mathrm{O}}$ instances from the unselected tasks, $\{1,\cdots,T\}\setminus t$.
    \STATE Adapt GMM parameters $\hat{\bm{\Theta}}(\mathcal{S};\bm{\Phi})$ to
    the support set by maximizing the likelihood with Eq.~(\ref{eq:gmm_params}).
   \STATE Calculate the OoD score by Eq.~(\ref{eq:u}) using the adapted GMM parameters.
    \STATE Calculate the smooth AUC by Eq.~(\ref{eq:smooth_auc}).
    \STATE Update common parameters $\bm{\Phi}$ using the smooth AUC and its gradient by a stochastic gradient method.
    \ENDWHILE
  \end{algorithmic}
\end{algorithm}

\subsection{Relation to existing methods}

A OoD detection method
based on Mahalanobis distance~\cite{lee2018simple}
assumes a GMM with a full covariance matrix shared by all classes,
\begin{align}
  p_{\mathrm{M}}\left(f(\vec{x};\bm{\phi})\mid\bm{\Theta}\right)
    = \sum_{k=1}^{K}\gamma_{k}\mathcal{N}\left(f(\vec{x};\bm{\phi})\mid
    \bm{\mu}_{k},\bm{\Sigma}\right).
    \label{eq:gmm_maha}
\end{align}
The negative OoD score is calculated by the minimum of
the Mahalanobis distance over classes,
\begin{align}
  u_{\mathrm{M}}(\vec{x}\mid\mathcal{S};\bm{\Phi})
  &= -\min_{k}[(f(\vec{x};\bm{\phi})-\hat{\bm{\mu}}_{k})^{\top}\hat{\bm{\Sigma}}^{-1}(f(\vec{x};\bm{\phi})-\hat{\bm{\mu}}_{k})]
  \nonumber\\
  &=-\max_{k}\log\mathcal{N}(f(\vec{x};\bm{\phi})\mid\hat{\bm{\mu}}_{k},\hat{\bm{\Sigma}})
  \nonumber\\
  &\approx-\log\sum_{k}\mathcal{N}(f(\vec{x};\bm{\phi})\mid\hat{\bm{\mu}}_{k},\hat{\bm{\Sigma}}),
\end{align}
where we used the fact that the LogSumExp is a smooth approximation
of the maximum, $\log\sum_{k}\exp(\cdot)\approx\max_{k}(\cdot)$.
Therefore, when class probability $\gamma_{k}$ are the same for all classes,
the Mahalanobis distance-based method can be seen that
it uses log likelihood in a latent space as the OoD score 
as with the proposed method. 

The prototypical network is a meta-learning method
for classification~\cite{snell2017prototypical},
and it assumes
the following
Gaussian mixture model with a spherical covariance in a latent space,
\begin{align}
  p_{\mathrm{P}}\left(f(\vec{x};\bm{\phi})\mid\bm{\Theta}\right)
    = \sum_{k=1}^{K}\gamma_{k}\mathcal{N}\left(f(\vec{x};\bm{\phi})\mid
    \bm{\mu}_{k},\tau\vec{I}\right),
    \label{eq:gmm_proto}
\end{align}
with uniform class probability $\gamma_{k}=\frac{1}{K}$.
Mean vectors $\{\bm{\mu}_{k}\}_{k=1}^{K}$ are obtained
by maximizing the log likelihood of the support set.
The proposed model can express more complicated distributions
by adapting a full covariance matrix for each class.
When parameter $\beta$ in Eq.~(\ref{eq:ml}) is large,
the proposed model becomes similar to a GMM with
a fixed spherical covariance as with the prototypical network.
By meta-learning $\beta$ such that the OoD detection performance is improved,
the proposed method can estimate the density flexibly
while avoiding overfitting.
Different from the proposed method,
the objective function of the outer optimization problem of
the prototypical network is the classification cross-entropy loss,
where the posterior class probability is calculated by
\begin{align}
  p(k\mid\vec{x})=\frac{\exp(-\frac{1}{2\tau}\parallel f(\vec{x};\bm{\phi})-\hat{\bm{\mu}}_{k}(\mathcal{S})\parallel^{2})}
  {\sum_{k'=1}^{K}\exp(-\frac{1}{2\tau}\parallel f(\vec{x};\bm{\phi})-\hat{\bm{\mu}}_{k'}(\mathcal{S})\parallel^{2})}.
  \label{eq:proto}
\end{align}

With classifier-based OoD detection methods~\cite{liang2018enhancing},
the negative OoD score is calculated by the maximum of the posterior class probability,
\begin{align}
  u_{\mathrm{P}}(\vec{x}\mid\mathcal{S};\bm{\Phi})&=-\max_{k} p(k\mid\vec{x})
  \propto -\max_{k} \mathcal{N}(f(\vec{x};\bm{\phi})\mid\hat{\bm{\mu}}_{k},\tau\vec{I}),
\end{align}
which is also related to the density estimation in a latent space.

Temperature scaling is used for OoD detection~\cite{liang2018enhancing},
where $\tau$ is
the temperature in Eq.~(\ref{eq:proto}),
and the temperature scaling controls the covariance.
Since the proposed method adapts the covariance for each class
to the support set,
the proposed method is related to the temperature scaling
that adapt to each task.

\section{Experiments}
\label{sec:experiments}

\subsection{Data}

We evaluated the proposed method
with the following six datasets:
Omniglot, Miniimagenet, CIFAR10, SVHN, Patent, and Dmoz.

The Omniglot dataset~\cite{lake2015human} consisted of
hand-written images of 964 characters from 50 alphabets (real and fictional).
There were 20 images for each character category.
Each image was represented by gray-scale with 28 $\times$ 28 pixels.
The number of instances, attributes, and categories were
19280, 784, and 964, respectively.

The Miniimagenet dataset
consisted of images from 100 categories~\cite{vinyals2016matching}.
Each image was represented by RGB with 84 $\times$ 84 pixels.
The number of instances, attributes, and categories were
60000, 21168, and 100, respectively.

The CIFAR10 dataset consisted of images from
10 categories~\cite{krizhevsky2009learning}.
Each image was represented by RGB with 32 $\times$ 32 pixels.
The number of instances, attributes, and categories were
60000, 3072, and 10, respectively.

The SVHN dataset consisted of digit images obtained from
house numbers in Google street view~\cite{netzer2011reading}.
Each image was represented by RGB with 32 $\times$ 32 pixels.
The number of instances, attributes, and categories were
73257, 3072, and 10, respectively.

The Patent dataset consisted of patents published
in Japan from January to March in 2004,
which were categorized by International Patent Classification.
Each patent was represented by bag-of-words.
We omitted words that occurred in fewer than 100 patents,
omitted patents with fewer than 100 unique words, and
omitted categories with fewer than ten patents.
The number of instances, attributes, and categories were
5714, 3201, and 285, respectively.

The Dmoz dataset consisted of webpages crawled in 2006
from the Open Directory Project~\cite{lorenzetti2019dmoz,lorenzetti2009semi}
\footnote{The data were obtained from \url{https://data.mendeley.com/datasets/9mpgz8z257/1}.}.
Each webpage was categorized in a web directory,
and represented by bag-of-words.
We omitted words that occurred in fewer than 300 webpages,
omitted webpages with fewer than 300 unique words,
and omitted categories with fewer than ten webpages.
The number of instances, attributes, and categories were
15159, 17659, and 354, respectively.

For each of the Omniglot, Miniimagenet, Patent, and Dmoz datasets,
we randomly used 60\% of the categories for training,
20\% for validation,
and the remaining categories for testing.
For each of the CIFAR10 and SVHN datasets,
since their number of classes was small,
the Miniimagenet dataset was used for the training and validation,
where we rescaled images in the Miniimagenet dataset
to 32 $\times$ 32 pixels.
For the test of the CIFAR10 dataset,
CIFAR10 images were used for the ID,
and SVHN images were used for the OoD.
For the test of the SVHN dataset,
SVHN images were used for the ID,
and CIFAR10 images were used for the OoD.
It is known that when a neural network-based density model
is trained on the CIFAR10 images,
the likelihood of the SVHN images is likely to be higher than that of
the CIFAR10 images~\cite{nalisnick2018deep}.
For all datasets, we generated 64 tasks for each validation and test data.
For each task in the validation and test data,
we first randomly selected five categories for ID,
and one category for OoD.
Then, we randomly selected five support instances
from each of the ID categories,
and five query instances from each of the ID and OoD categories.
We performed five experiments with different data splits
for each dataset.

\subsection{Comparing methods}

We compared our proposed method (Ours) with
ablation of the proposed method (Ours-C),
three types of prototypical networks~\cite{snell2017prototypical}
(Proto-C, Proto-A, and Proto-O),
three types of model-agnostic meta-learning~\cite{finn2017model}
(MAML-C, MAML-A, and MAML-O),
OoD-MAML~\cite{jeong2020ood}, 
learnable class boundary networks~\cite{wang2019out} (LCBO),
Mahalanobis distance-based method~\cite{lee2018simple} (Mahalanobis),
deep support vector data description (SVDD)~\cite{ruff2018deep},
density estimation with a single Gaussian model (Gaussian),
and kernel density estimation (KDE) in a latent space.
All methods use neural networks
to map instances from the original space to a latent space.
`C' in the method names indicates that
the method is meta-learned 
by minimizing the expected test classification cross-entropy loss,
`A' indicates that
the method is meta-learned
by maximizing the AUC of OoD detection,
and `O' indicates that
the method is based on
out-of-distribution detector for neural networks
(ODIN)~\cite{liang2018enhancing}.

The methods based on prototypical networks
(Proto-C, Proto-A, and Proto-O)
assume a Gaussian mixture model
with a unit spherical covariance in a latent space,
and the class probability is calculated based o the Bayes rule.
The methods based on model-agnostic meta-learning~\cite{finn2017model}
(MAML-C, MAML-A, and MAML-O)
model the class probability by a softmaxed linear projection
from the latent space.
With the methods based on ODIN (Proto-O and MAML-O),
the temperature of the class probability is meta-learned,
and small perturbations are added to the input
to the direction that maximizes the maximum class probability.
The neural network is meta-learned
by minimizing the expected test classification cross-entropy loss.
With prototypical network-based and MAML-based methods,
the negative OoD score is calculated by
the maximum of the class probability,
$\max_{k}p(k\mid\vec{x})$.
With OoD-MAML,
fake OoD instances are generated with the framework of MAML.
With learnable class boundary networks (LCBO),
the OoD score is modeled by
a neural network that takes the mean for each class and
the instance in the latent space as input.
With Mahalanobis,
the negative OoD score is calculated by
the minimum Mahalanobis distance to the class mean in the latent space.
SVDD, Gaussian, and KDE do not use class label information
in the support set.
With SVDD~\cite{ruff2018deep},
the OoD score is calculated by the distance from
the mean of the support instances in the latent space.
With Gaussian and KDE,
the OoD score is calculated by the likelihood in the latent space.
Except for Ours-C, Proto-C, MAML-C, Proto-O, and MAML-O,
the neural network is meta-learned
by maximizing the AUC of OoD detection.

\subsection{Setting}

In image datasets (Omniglot, Miniimage, CIFAR10, and SVHN),
we used a four-layered convolutional neural network of
filter size 32, kernel size three, and padding size one for $f$
with all methods.
In text datasets (Patent and Dmoz),
we used a three-layered feed-forward neural network
with 256 hidden and output units for $f$.
The instances in a latent space
were normalized by dividing by the standard deviation
of the support set.
For the activation function,
we used rectified linear unit $\mathrm{ReLU}(x)=\max(0,x)$.
We optimized using Adam~\cite{kingma2014adam}
with learning rate $10^{-3}$,
and dropout rate $10^{-1}$~\cite{srivastava2014dropout}.
For MAML-based method, we used three inner epochs.
The validation data were used for early stopping,
for which the maximum number of training epochs was 5,000.
Our implementation was based on PyTorch~\cite{paszke2017automatic}.

\subsection{Results}

\begin{table}[t]
  \centering
  \caption{Average test AUC for OoD detection and its standard error. Values in bold are not statistically significantly different at the 5\% level from the best performing method in each column according to a paired t-test.}
  \label{tab:auc}
  {\tabcolsep=0.4em
    \begin{tabular}{lrrrrrr}
      \hline
      & Omniglot & Miniimagenet & CIFAR10 & SVHN & Patent & Dmoz \\
      \hline
      Ours &{\bf 0.992$\pm$0.001} &{\bf 0.673$\pm$0.005} &0.619$\pm$0.017 &{\bf 0.989$\pm$0.001} &{\bf 0.873$\pm$0.006} &{\bf 0.828$\pm$0.005} \\
 Ours-C & 0.990$\pm$0.001 & 0.621$\pm$0.006 & 0.471$\pm$0.034 & {\bf 0.989$\pm$0.000} & 0.824$\pm$0.009 & 0.785$\pm$0.006 \\      
 Proto-C & 0.888$\pm$0.007 & 0.577$\pm$0.006 & 0.565$\pm$0.012 & 0.369$\pm$0.008 & 0.737$\pm$0.009 & 0.705$\pm$0.006 \\
 Proto-A & 0.980$\pm$0.002 & 0.185$\pm$0.014 & 0.351$\pm$0.022 & 0.274$\pm$0.034 & 0.827$\pm$0.010 & 0.783$\pm$0.006 \\
 Proto-O & 0.768$\pm$0.012 & 0.567$\pm$0.005 & {\bf 0.799$\pm$0.010} & 0.199$\pm$0.013 & 0.163$\pm$0.012 & 0.072$\pm$0.005 \\
 MAML-C & 0.935$\pm$0.004 & 0.592$\pm$0.003 & 0.605$\pm$0.007 & 0.350$\pm$0.013 & 0.623$\pm$0.010 & 0.563$\pm$0.005 \\
 MAML-A & 0.521$\pm$0.009 & 0.498$\pm$0.006 & 0.636$\pm$0.022 & 0.309$\pm$0.040 & 0.487$\pm$0.014 & 0.464$\pm$0.005 \\
 MAML-O & 0.498$\pm$0.011 & 0.504$\pm$0.005 & 0.634$\pm$0.023 & 0.466$\pm$0.010 & 0.511$\pm$0.008 & 0.003$\pm$0.001 \\
 OoD-MAML & 0.982$\pm$0.003 & {\bf 0.672$\pm$0.003} & 0.669$\pm$0.025 & 0.656$\pm$0.024 & 0.810$\pm$0.006 & 0.767$\pm$0.007 \\
 LCBO & 0.559$\pm$0.005 & 0.522$\pm$0.007 & 0.481$\pm$0.012 & 0.541$\pm$0.032 & 0.573$\pm$0.008 & 0.564$\pm$0.006 \\
 Mahalanobis & 0.989$\pm$0.002 & 0.520$\pm$0.006 & 0.623$\pm$0.011 & {\bf 0.988$\pm$0.001} & {\bf 0.875$\pm$0.007} & {\bf 0.826$\pm$0.007} \\
 SVDD & 0.832$\pm$0.003 & 0.528$\pm$0.003 & 0.462$\pm$0.033 & {\bf 0.718$\pm$0.050} & 0.693$\pm$0.009 & 0.619$\pm$0.005 \\
 Gauss & 0.989$\pm$0.000 & 0.539$\pm$0.002 & 0.626$\pm$0.011 & 0.984$\pm$0.002 & {\bf 0.878$\pm$0.007} & {\bf 0.828$\pm$0.007} \\
 KDE & 0.858$\pm$0.006 & 0.511$\pm$0.015 & 0.421$\pm$0.029 & 0.706$\pm$0.032 & 0.733$\pm$0.008 & 0.655$\pm$0.016 \\
\hline
\end{tabular}}
\end{table}

\begin{figure*}[t!]
  \centering
  {\tabcolsep=-0.8em
    \begin{tabular}{ccc}
      \includegraphics[width=16em]{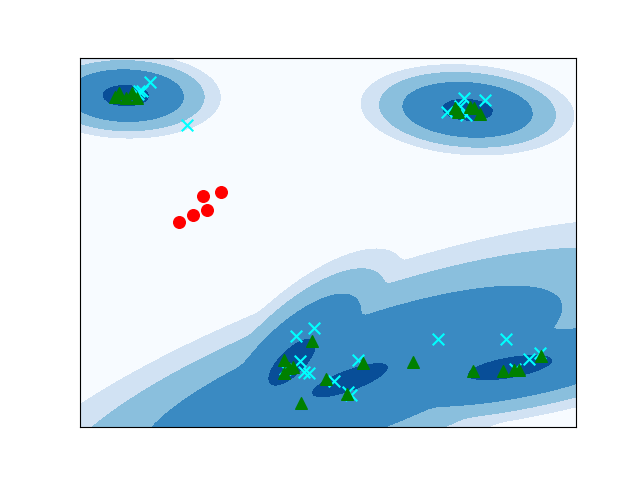}&
      \includegraphics[width=16em]{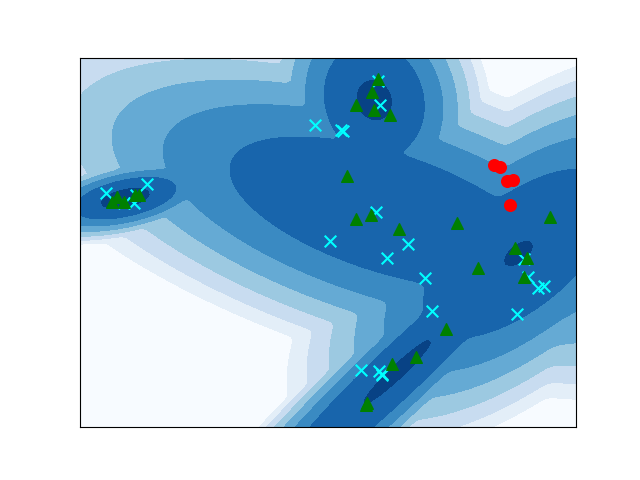}&
      \includegraphics[width=16em]{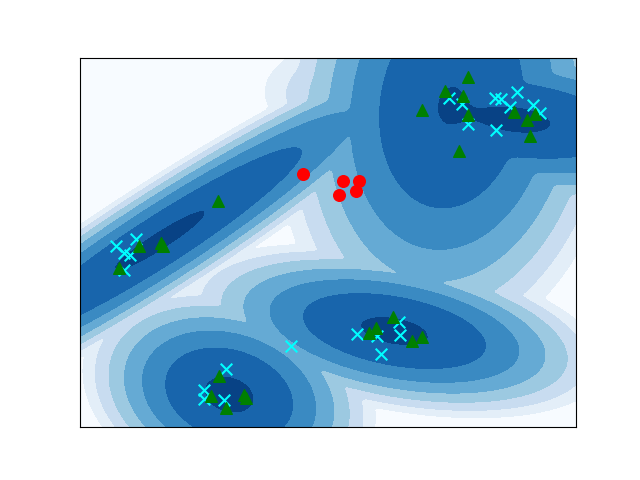}\\
  \end{tabular}}
  \caption{Examples of density estimation in a two-dimensional latent space by the proposed method on the Omniglot dataset. The green triangles are support instances, blue crosses are ID query instances, and red circles are OoD query instances. The contours represent the log likelihood, i.e., negative OoD score,
    where the darker color indicates the higher log likelihood.}
  \label{fig:vis}
\end{figure*}

Table~\ref{tab:auc} shows the test AUC.
Our proposed method achieved high AUC with all datasets.
With the CIFAR10 dataset, the AUC by the proposed method
was lower than that by Proto-O.
However, the AUC by Proto-O with the SVHN dataset was very low.
It indicates that Proto-O estimated that the CIFAR10 images were ID and
the SVHN images were OoD by meta-learning with the Miniimagenet dataset.
By meta-learning a neural network for density estimation in a latent space,
the proposed method achieves a relatively high AUC on
both of the CIFAR10 and SVHN datasets.

The proposed method meta-trained by maximizing
the OoD detection performance (Ours)
was better than that by minimizing the classification loss (Ours-C).
This result indicates that the
directly maximizing the OoD detection performance is important.

The prototypical network-based methods performed worse
than the proposed method.
This result demonstrates the effectiveness of density estimation by
full covariance GMMs in a latent space.
The MAML-based methods adapt the whole neural network to the support set.
On the other hand, the proposed method adapts only a GMM in a latent space to the support set.
The better performance of the proposed method than the MAML-based methods
indicates that the GMM in a latent space is effective for OoD detection
when the number of instances in the support set is small.
OoD-MAML improved the AUC compared with the other MAML-based methods
by generating fake OoD instances.
However, since it is difficult to generate
a wide variety of OoD instances for each task,
it was worse than the proposed method.
The proposed method detects OoD by density estimation
using only ID instances without OoD instances,
and it does not need to generate fake OoD instances.

With LCBO, the OoD score is calculated by a neural network
that takes the support set as input.
Therefore, the OoD score of the support set is not necessarily high.
In contrast, with the proposed method,
the density-based OoD score of the support set is high
since we maximize the likelihood of the support set.
Therefore, the proposed method can flexibly adapt to a given support set
even when it does not appear in the meta-training dataset,
and it led to better performance of the proposed method.
While Mahalanobis achieved good performance,
it was worse than the proposed method on the Omiglot and Miniimagenet datasets.
This result indicates the effectiveness to model
class-specific full covariances in GMMs.
The AUC by the methods that do not use class label information
was lower than the proposed method.
This result shows the usefulness of the class label information
for OoD detection.

Figure~\ref{fig:vis} shows a visualization of density estimation in a two-dimensional latent space by the proposed method on the Omniglot dataset.
We additionally used a three-layered feed-forward neural network
for the two-dimensional embedding after the convolutional neural network.
By the neural network, ID instances were embedded in a high-density area,
and OoD instances were embedded in a low-density area.
By the GMMs, the task-specific density function was modeled flexibly
in the latent space depending on the support set.

\begin{figure}[t!]
  \centering
  \includegraphics[width=19em]{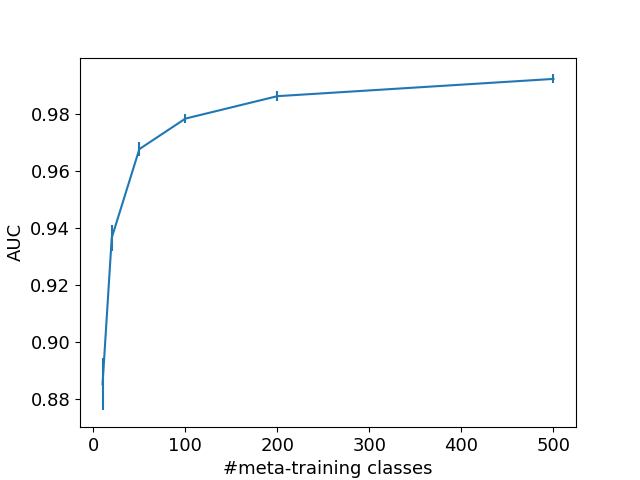}
  \caption{Average test AUC for OoD detection with different numbers of meta-training classes by the proposed method on the Omniglot dataset. The bar shows standard error.}
  \label{fig:auc_class}
\end{figure}

\begin{figure}[t!]
  \centering
  \includegraphics[width=19em]{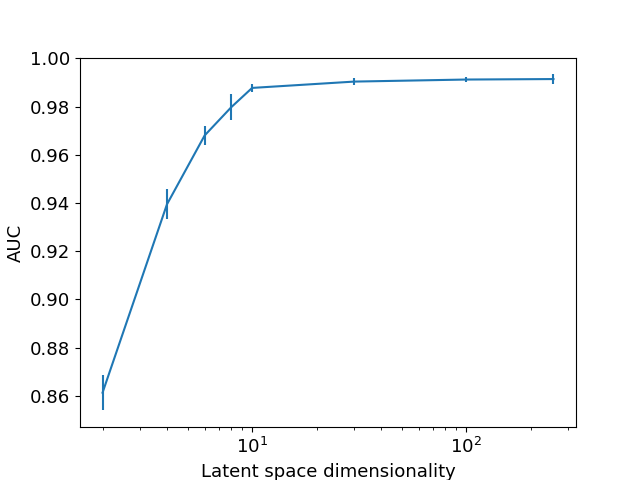}
  \caption{Average test AUC for OoD detection with different latent space dimensionality by the proposed method on the Omniglot dataset. The bar shows standard error.}
  \label{fig:auc_dim}
\end{figure}

\begin{table}[t]
  \centering
  \caption{Average test classification accuracy of class labels and its standard error. Values in bold are not statistically significantly different at the 5\% level from the best performing method in each column according to a paired t-test. Since LCBO, SVDD, Gauss, and KDE did not output the class label, we omit them.}
  \label{tab:acc}
  {\tabcolsep=0.4em
    \begin{tabular}{lrrrrrr}
      \hline
      & Omniglot & Miniimagenet & CIFAR10 & SVHN & Patent & Dmoz \\
      \hline
Ours &{\bf 0.995$\pm$0.000} &0.555$\pm$0.006 &{\bf 0.497$\pm$0.006} &{\bf 0.276$\pm$0.006} &0.880$\pm$0.003 &0.816$\pm$0.005 \\
 Ours-C & {\bf 0.994$\pm$0.001} & {\bf 0.602$\pm$0.005} & {\bf 0.505$\pm$0.007} & {\bf 0.269$\pm$0.004} & 0.868$\pm$0.004 & {\bf 0.822$\pm$0.001} \\
 Proto-C & {\bf 0.994$\pm$0.000} & {\bf 0.590$\pm$0.004} & {\bf 0.489$\pm$0.010} & {\bf 0.270$\pm$0.004} & 0.887$\pm$0.002 & 0.823$\pm$0.005 \\
 Proto-A & {\bf 0.995$\pm$0.000} & 0.293$\pm$0.012 & 0.267$\pm$0.009 & 0.198$\pm$0.002 & 0.879$\pm$0.001 & {\bf 0.825$\pm$0.004} \\
 Proto-O & 0.991$\pm$0.001 & {\bf 0.589$\pm$0.006} & 0.475$\pm$0.009 & 0.251$\pm$0.008 & 0.883$\pm$0.002 & 0.821$\pm$0.003 \\
 MAML-C & 0.990$\pm$0.001 & 0.576$\pm$0.006 & 0.489$\pm$0.006 & 0.239$\pm$0.003 & 0.754$\pm$0.005 & 0.692$\pm$0.007 \\
 MAML-A & 0.197$\pm$0.003 & 0.200$\pm$0.001 & 0.198$\pm$0.002 & 0.201$\pm$0.001 & 0.205$\pm$0.004 & 0.208$\pm$0.004 \\
 MAML-O & 0.678$\pm$0.071 & 0.522$\pm$0.008 & 0.433$\pm$0.020 & 0.219$\pm$0.004 & 0.239$\pm$0.016 & 0.658$\pm$0.020 \\
 OoD-MAML & 0.988$\pm$0.002 & 0.563$\pm$0.010 & 0.443$\pm$0.007 & 0.223$\pm$0.004 & 0.765$\pm$0.006 & 0.724$\pm$0.008 \\
 Mahalanobis & {\bf 0.994$\pm$0.001} & 0.319$\pm$0.005 & 0.491$\pm$0.007 & {\bf 0.275$\pm$0.006} & {\bf 0.905$\pm$0.002} & {\bf 0.839$\pm$0.006} \\
 \hline
\end{tabular}}
\end{table}

Figure~\ref{fig:auc_class} shows the test AUC with different numbers 
of meta-training classes by the proposed method on the Omniglot dataset.
The test AUC increased as the number of meta-training classes increased.
This result indicates that the proposed method can increase the performance
by collecting meta-training data from many classes.

Figure~\ref{fig:auc_dim} shows the test AUC with different latent space
dimensionality by the proposed method on the Omniglot dataset.
When the latent space dimensionality was small, the test AUC was low.
It is because in a low-dimensional space representing complicated instances is difficult, and the density estimated with GMMs does not have enough expressive power.

Table~\ref{tab:acc} shows the test classification accuracy of class labels.
Since the proposed method (Ours)
is not meta-learned to improve the classification
performance, it was worse than the proposed method that is meta-learned
by minimizing the classification loss (Ours-C).
Since we focus on improving OoD detection performance in this paper,
we use the AUC on OoD detection as the objective function.
When we also want to improve classification performance,
we can add the classification cross-entropy loss in the objective function.
The classification accuracy by Ours-C was almost the same with
that by Proto-C.
In contrast, the AUC for OoD detection by Ours was better than Proto-A.
This result indicates that 
although
GMMs with a common spherical covariance are sufficient for classification,
GMMs with full covariance for each class are needed for OoD detection.

Table~\ref{tab:time}
shows the average computational time for meta-training
and meta-testing
with a GTX 1080Ti GPU. The computational
time for the MAML-based methods was much longer than the other methods.
It was because MAML-based methods required
iterative gradient descent steps for the inner optimization.
Also, the ODIN-based methods took a long time since
they required to perturb the input for the query instances.

\begin{table}[t!]
  \centering
  \caption{Average meta-training and meta-testing computational time in seconds on the Omniglot dataset.}
  \label{tab:time}
\begin{tabular}{lrr}
\hline
& Train & Test \\
\hline
Ours &16619.6 & 0.416 \\
 Ours-C & 15230.5 & 0.329 \\
 Proto-C & 13381.7 & 0.325 \\
 Proto-A & 12846.6 & 0.345 \\
 Proto-O & 25299.8 &0.827 \\
 MAML-C & 57677.4 & 0.960 \\
 MAML-A & 55057.0 & 1.191 \\
 MAML-O & 177137.1 & 1.727 \\
 OoD-MAML & 47581.7 & 0.928 \\
 LCBO & 10673.2 & 0.198 \\
 Mahalanobis & 15169.2 & 0.283 \\
 SVDD & 10370.6 & 0.259 \\
 Gauss & 15056.3 & 0.294 \\
 KDE & 11922.2& 0.299 \\
\hline
\end{tabular}
\end{table}

\section{Conclusion}
\label{sec:conclusion}

We proposed a meta-learning method for OoD detection.
Our proposed method trains a neural network such that
density-based OoD scores perform well when
a Gaussian mixture model in the latent space is adapted
to given in-distribution data for each task.
Experiments on six datasets confirmed
that our proposed method had better OoD detection performance
than existing methods did.
For future work,
we will use other density estimation methods in a latent space
in our framework.
Also, we want to improve the performance by additionally using
OoD techniques such as input perturbations
and the use of representations in multiple layers~\cite{lee2018simple}.

\bibliography{arxiv_ood}
\bibliographystyle{abbrv}

\end{document}